%% file: main.tex
\pdfoutput=1

\documentclass[11pt]{article}

\usepackage[]{acl}

\usepackage{times}
\usepackage{latexsym}

\usepackage[T1]{fontenc}

\usepackage[utf8]{inputenc}

\usepackage{microtype}

\usepackage{inconsolata}

\usepackage{graphicx}
\usepackage{stmaryrd}
\usepackage{amsfonts}
\usepackage{booktabs,arydshln}
\usepackage{multirow}
\usepackage{graphicx}
\usepackage{url}
\usepackage{hyperref}
\usepackage{wrapfig}
\usepackage{amsmath}
\usepackage{bbm}
\usepackage{algorithm}
\usepackage{algorithmicx}
\usepackage[noend]{algpseudocode}
\usepackage{epigraph}
\usepackage{amssymb}
\usepackage{amsthm}
\usepackage{tcolorbox}
\usepackage{soul}
\usepackage{bbm}
\usepackage{xcolor}
\definecolor{DarkGreen}{RGB}{30,130,30}
\usepackage{makecell}
\usepackage{pifont}
\usepackage{listings}

\newcommand{\cmark}{\textcolor{DarkGreen}{\ding{51}}}
\newcommand{\xmark}{\textcolor{red}{\ding{55}}}

\newcommand*\Let[2]{\State #1 $\gets$ #2}
\algrenewcommand\algorithmicrequire{\textbf{Input:}}
\algrenewcommand\algorithmicensure{\textbf{Parameters:}}
\algnewcommand{\parState}[1]{
  \parbox[t]{\dimexpr\linewidth-\algmargin}{\strut #1\strut}}

\newcommand{\Sref}[1]{\S\ref{#1}}
\newcommand{\method}{\textsc{BloomScrub}}

\newcommand{\jack}[1]{{\color{orange}[JZ: #1]}}

\definecolor{darkgreen}{RGB}{0,140,0}

\newcommand{\Mycomment}[1]{\Comment{\textcolor{darkgreen}{#1}}}

\title{Certified Mitigation of Worst-Case LLM Copyright Infringement}

\newcommand{\aspace}{\hspace{1.05em}}

\author{%
    Jingyu Zhang\aspace
    Jiacan Yu\aspace
    Marc Marone\aspace
    Benjamin Van Durme\aspace
    Daniel Khashabi\\
    Johns Hopkins University \\ 
    \texttt{\{jzhan237,jyu197,mmarone1\}@jhu.edu}
}

\begin{document}
\maketitle
\begin{abstract}

The exposure of large language models (LLMs) to copyrighted material during pre-training raises concerns about unintentional copyright infringement post deployment. This has driven the development of ``copyright takedown'' methods---post-training approaches aimed at preventing models from generating content substantially similar to copyrighted ones. 
While current mitigation approaches are somewhat effective for \textit{average-case} risks, we demonstrate that they overlook \textit{worst-case} copyright risks exhibited by the existence of long, verbatim quotes from copyrighted sources.  
We propose \method, a remarkably simple yet highly effective inference-time approach that provides \textit{certified} copyright takedown.
Our method repeatedly interleaves quote detection with rewriting techniques to transform potentially infringing segments.
By leveraging efficient data sketches (Bloom filters), our approach enables scalable copyright screening---even for large-scale real-world corpora.
When quotes beyond a length threshold cannot be removed, the system can abstain from responding, offering certified risk reduction. 
Experimental results show that \method{} reduces infringement risk, preserves utility, and accommodates different levels of enforcement stringency with adaptive abstention.
Our results suggest that lightweight, inference-time methods can be surprisingly effective for copyright prevention.

\end{abstract}

\section{Introduction}
 
\begin{figure*}[t]
    \centering
    \includegraphics[width=\linewidth]{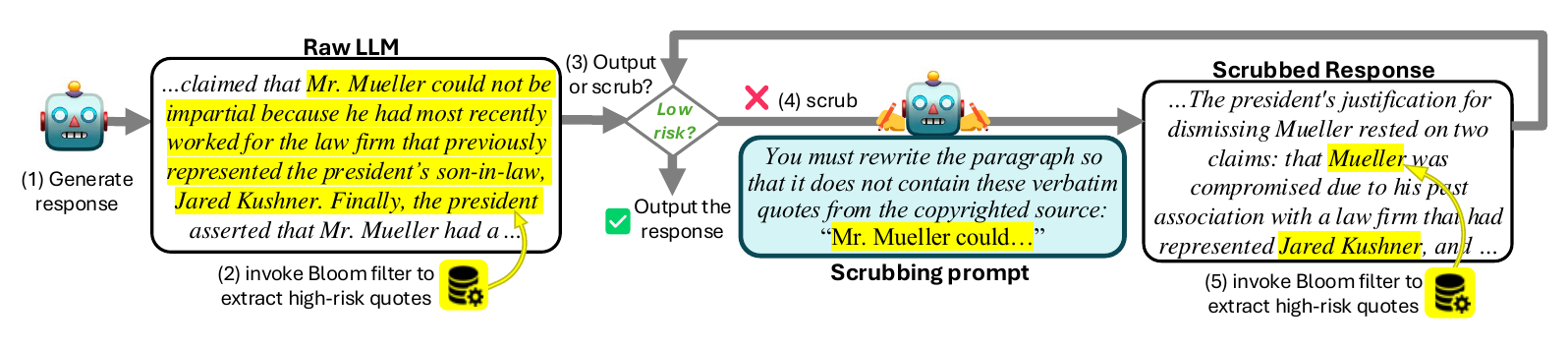}
    \caption{\method{} works by interleaving two key steps: (1) using a Bloom filter to extract high-risk quotes from model responses, and (2) apply guided rewriting to ``scrub'' these quotes from the text. This iterative process ensures removal of high-risk quotes while preserving utility.}
    \label{fig:teaser}
\end{figure*}

Large language models (LLMs) are trained on vast datasets, many of which include copyrighted material or content with usage restrictions~\citep[][\textit{i.a.}]{bandy2021addressingdocumentationdebtmachine, https://doi.org/10.1111/jwip.12331}. This raises legal and ethical concerns, particularly regarding unauthorized reproduction of copyrighted content in model outputs. In the U.S., model creators often invoke the \textit{fair use} doctrine---a legal defense established long before the rise of LLMs—that permits the use of copyrighted data for training under certain conditions, typically based on factors like purpose, scope, and market impact~\cite{lemley2020fair}.  

However, the boundaries of fair use in AI remain uncertain, as courts and regulators struggle to keep up with the rapid evolution of LLMs. The greatest legal risk arises when a model outputs content that is substantially similar to copyrighted material---particularly long verbatim excerpts---which weakens a fair use defense and increases the likelihood of legal challenges~\cite{henderson2023foundation}.  
A notable example is the New York Post lawsuit against Perplexity AI, which alleges that the company engaged in “massive illegal copying”, reproducing copyrighted content without authorization~\cite{dowjones2024}. Cases like this underscore a critical point: preventing long verbatim quotations from copyrighted sources is essential in mitigating copyright risk. While this alone may not be a comprehensive safeguard, it is a crucial first step in ensuring transformative use.

In this work, we aim to prevent models from generating long, sensitive quoted statements originating from copyrighted documents, which we term as the \textit{worst-case} risk of copyright infringement. 
Although this might seem straightforward, existing copyright prevention methods fail to fully eliminate problematic content or do so at the cost of severely degrading text utility. As our empirical results (\Sref{sec:exp}) show, current mitigation techniques leave LLMs vulnerable to legal liability by failing to reliably prevent long verbatim outputs.

To address this gap, we propose \method~(Fig.~\ref{fig:teaser}), a remarkably simple yet highly effective inference-time approach that provides \textit{certified copyright takedown}---completely eliminating long verbatim quotes from copyrighted sources---for large-scale corpora while preserving text quality. Shown in Fig.~\ref{fig:teaser}, \method{} operates in two alternating steps:  (1) Quoted span detection, which employs a Bloom filter~\citep{bloom1970spacetime} to efficiently identify verbatim segments at scale, even against massive copyrighted corpora. (2) Dynamic rewriting (``scrubbing'') mechanism, which diffuses high-risk detected quotes, ensuring compliance with copyright constraints while maintaining fluency and coherence. These steps are repeated iteratively until the output is classified as low-risk, i.e., the length of the longest quoted span falls within a predefined risk threshold. If a response cannot achieve low-risk status within a limited number of iterations, the system opts to abstain from generating an answer (\Sref{subsec:our:approach}).

Despite its simplicity, \method{} offers key advantages. It is \textbf{scalable}, with Bloom filters enabling efficient large-scale corpus screening for real-world deployment. It is \textbf{plug-and-play}, allowing users to easily update the targeted copyrighted corpus by integrating it into the Bloom filter sketch. It is \textbf{adaptive}, as the rewriting mechanism dynamically adjusts to different levels of copyright enforcement for precise risk mitigation. Finally, it is \textbf{certified}, formally guaranteeing the removal of long verbatim quotes and abstaining from generating responses when compliance cannot be ensured.

Our experimental results demonstrate that, compared to existing methods such as MemFree Decoding~\citep{ippolito2022preventing} and Reversed Context-Aware Decoding~\citep{shi2023trusting, wei2024evaluatingcopyrighttakedownmethods}, which modifies the autoregressive decoding process to steer models away from generating copyrighted content, \method{} is both more effective at mitigating copyright risks and more flexible in preserving text utility. Furthermore, \method{} allows dynamic adjustment of risk thresholds by varying the number of rewrite iterations, offering a scalable and adaptive solution. Finally, we analyze the failure modes of prior approaches and demonstrate how \method{} overcomes these limitations, providing a practical and robust framework for certified copyright takedown in deployed LLMs.

In summary, our contributions are: 
(1) We introduce the task of certified copyright takedown, focusing on the worst-case risk of long verbatim quotes from copyrighted sources.  
(2) We propose \method, an efficient, inference-time solution using Bloom filters and dynamic rewriting for scalable copyright prevention.  
(3) We empirically demonstrate that \method{} outperforms existing methods in both risk mitigation and utility preservation.  

\begin{table*}[ht]
\centering
\resizebox{\linewidth}{!}{
\begin{tabular}{@{}lccccc@{}}
\toprule
Property $\shortdownarrow$ - Approach $\shortrightarrow$ & \makecell{Unlearning\\{\scriptsize (Varivous works)}} & \makecell{SysPrompt\\{\scriptsize (Varivous works)}} & \makecell{MemFree\\{\scriptsize\citep{ippolito2022preventing}}} & \makecell{R-CAD\\{\scriptsize\citep{wei2024evaluatingcopyrighttakedownmethods}}} &  \makecell{\method\\(Ours)} \\ \midrule
Avoids quoting from $C$? & \cmark & \cmark & \cmark & \cmark & \cmark \\
Retains the knowledge in $C$? & \xmark & \cmark & \cmark & \cmark & \cmark \\
Doesn't require model to support system prompt? & \cmark & \xmark & \cmark & \cmark & \cmark \\
Operates without access to the model logits? & \xmark & \cmark & \xmark & \xmark & \cmark \\
Works without \textit{\textbf{direct}} access to $C$ during mitigation? & \xmark & \cmark & \cmark & \xmark & \cmark \\ \bottomrule
\end{tabular}
}

\caption{Comparisons of common copyright mitigation approaches. $C$ denotes a large-scale copyrighted corpus. Our \method{} is the most lightweight and plug-and-play of the methods considered, applicable to a wide range of settings without requiring logits nor direct access to $C$, since only a Bloom filter representation of $C$ is needed.
}
\label{tab:setup}
\end{table*}

\section{Background and Related Work}

\paragraph{Memorization in LLMs}
Contemporary LLMs are shown to have memorized portions of their training data~\citep{carlini21extracting, carlini2022quantifying, hu2022membership, NEURIPS2023_59404fb8, Hartmann2023SoKMI}, and can regurgitate verbatim copies of copyrighted material~\citep{karamolegkou2023copyrightviolationslargelanguage, chang2023speak, lee2023talkin, meeus2024copyrighttrapslargelanguage}.
These works establish that memorization is an ongoing risk with models, for both quality \cite{lee2022deduplicating} and impermissible copying.

\paragraph{Fair Use}
In the US, despite the existence of the fair use doctrine~\citep[17 U.S.C. §107;][]{lemley2020fair}, current LLMs are still at risk for copyright disputes since substantially similar content---such as \textit{long} \textit{verbatim} quotes of copyrighted material---is often out of scope of fair use. Motivated by \textit{transformativeness} as a key aspect of fair use, \citet{henderson2023foundation} encourage research into ``technical mitigations'' around transformations of both low-level and high-level content, noting that ``low-level'' content can involve n-gram overlap.
\citet{wei2024evaluatingcopyrighttakedownmethods} recently proposed the notion of {copyright takedown} for ensuring models do not generate content substantially similar to copyrighted material while preserving utility. 
Relatedly, \citet{chen2024copybenchmeasuringliteralnonliteral} measure both literal and non-literal copying in the domain of fiction books.
The landscape around LLMs and fair use is rapidly developing, but these recent works highlight that current LLMs are at risk of copyright violations unless actively mitigated.

\paragraph{Mitigation approaches}

A popular thread of work focus on adapting ``unlearning'' for the goal of copyright mitigation~\citep{eldan2023whosharrypotterapproximate, hans2024likegoldfishdontmemorize, maini2024tofutaskfictitiousunlearning, dou2024avoidingcopyrightinfringementlarge}. 
However, because the original intended goal of unlearning is  \textit{forgetting} (i.e, forget a given dataset $\mathcal{D}$ as if the model has not been trained on $\mathcal{D}$), this is undesirable for copyright purposes due
to its high risk for utility loss, i.e., the failure to preserve uncopyrightable factual knowledge ~\citep{wei2024evaluatingcopyrighttakedownmethods}. 
In the United States, common factual knowledge contained within copyrighted material is generally not copyrightable, as established by the landmark \citet{feist1991feist} case, though compilations of facts may receive protection if they exhibit an original selection or arrangement. Consequently, complete forgetting is an {overkill} in many practical settings. \citet{liu2024shieldevaluationdefensestrategies} propose an agent-based copyright defense mechanism by utilizing web services to verify copyright status of prompts. Other inference-time copyright mitigation approaches such as incorporating system prompt~\citep{wei2024evaluatingcopyrighttakedownmethods, chen2024copybenchmeasuringliteralnonliteral} or blocking $n$-grams from copyrighted corpus through MemFree decoding~\citep{ippolito2022preventing} better preserves information in copyrighted content but are at risk of infringement in the worse case, as shown by our results in \Sref{sec:exp}. We bridge this gap on \textit{worst-case} infringement by proposing \method, an inference-time takedown method that is scalable, effective, and certified.

\section{A Certified Copyright Protection Approach}

We first define the task and our metrics for assessing the generation of quotes from copyrighted sources (\S\ref{subsec:task-definition}).
We then define our algorithm for for dynamic rewriting and show that it is effective and flexible compared to other methods (\S\ref{subsec:our:approach}).

\subsection{Formulating the Task of Removing Long Verbatim Quotes} 
\label{subsec:task-definition}
Key aspects of \textit{Fair Use} include \textbf{transformativeness} and the \textbf{amount} of content \citep[17 U.S.C. §107;][]{henderson2023foundation}. It is therefore desirable for LLMs to avoid generating long verbatim quotes from copyrighted sources, even while the use of the underlying knowledge may be permitted under fair use. 

Given a corpus $C$, the goal of the certified copyright takedown task is preventing verbatim quotes from $C$ being generated. We assume a tolerance $\tau$, where \underline{any} verbatim match of text $y$ with length $|y| > \tau$ is considered risky. We measure the \textbf{\textit{worst-case infringement}} outcome and propose a novel metric $\%R>Q(\tau)$ (\Sref{sec:exp_setup}) to facilitate evaluation of matched quotes over massive corpora, while also \textbf{\textit{average-case infringement}}. 
In addition to low similarity with copyrighted documents, LLMs should be able to preserve uncopyrightable information and factual knowledge in copyright data. Thus, we also measure the \textit{\textbf{information quality}} of generated responses and \textit{\textbf{utility}} measured by fact-related QA. We defer further details to \Sref{sec:exp_setup}.

The total elimination of long quotes might lead to {overprotection}, e.g., certain named entities or phrases can exceed the threshold $\tau$ while being perfectly reasonable to quote. We discuss this in our analysis (\Sref{sec:long_quotes_analysis}) and find that the adaptive LLM-based rewriting of \method{} can serve as a ``soft removal'' mechanism, and preserve these named entities when rewriting is infeasible. In contrast, MemFree decoding's hard removal approach always prevents long-enough $n$-grams from being generated~\citep{ippolito2022preventing}, causing greater quality and utility loss.

\subsection{\method: Dynamic Guided Rewrite for Copyright Takedown}
\label{subsec:our:approach}

We now introduce \method, a plug-and-play approach for dynamic guided rewriting to mitigate copyright risks. Shown in Table~\ref{tab:setup}, \method{} requires only black-box access to the generation model and operates by dynamically detecting copyrighted quotes using signals from a Bloom filter. When a rewrite is necessary, \method{} identifies verbatim quotes that must be modified and invokes a rewrite model to reduce copyright risk.

Our method first detects copied quotes and then rewrites the content to avoid overlap. It also triggers an \textit{abstention} in the event that the amount of copying cannot be reduced below a certain threshold.
These steps do not ensure total compliance, but are a step towards better mitigation.

\input{alg}

\paragraph{(A) Fixed-width Bloom filter for quote extraction}

We first detail the quote extractor component of \method. 
Given a large-scale corpus $C$ containing copyrighted content (which we want to avoid regurgitating) and a generated response $y$, we use a Bloom filter to extract substrings of $y$ that is verbatim quoted from $C$. Specifically, given granularity $n$, we use Data Portraits~\citep{marone2023dataportraits} to index all character $n$-grams in $C$ into a Bloom filter.\footnote{We conduct normalization of whitespaces, punctuations, and cases.} The quote extractor $\mathcal{E}_C$ is implemented by querying each $n$-gram of $y$ to the Bloom filter and checking for hits. When $k$ \textit{continuous hits} of multiple $n$-grams with 1 character offset is detected, $\mathcal{E}_C$ aggregate them into a single long quote of length $n+k-1$.\footnote{For example, if \texttt{abcd}, \texttt{bcde} and \texttt{cdef} are hits, they are aggregated into a single quote, \texttt{abcdef}.} This mechanism will merge sufficiently overlapped short quotes into a single longer one, allowing the detection of near-verbatim ``stitched quotes'' which also contributes to copyright risks~\citep{chen2024copybenchmeasuringliteralnonliteral}. Because Bloom filter's zero false negative property~\citep{bloom1970spacetime}, \emph{all quotes of length at least $n$ is guaranteed to be extracted, providing certification of the extraction of long quotes.}\footnote{This is because for a quote $q=c_1\dots c_k$ of length $k\geq n$, every $n$-gram substring of $q$, $c_1\dots c_n, c_2\dots c_{n+1}, \dots$ are guaranteed to be matched. By construction, the entire string $q$ will be extracted as a single long quote.} We set $n=25$ in our experiments, a relatively small number, to ensure coverage of  all quotes that are at risk of infringement.

\begin{figure*}[t]
    \centering
    \includegraphics[width=\linewidth,trim=0cm 0.3cm 0cm 0cm,clip=true]{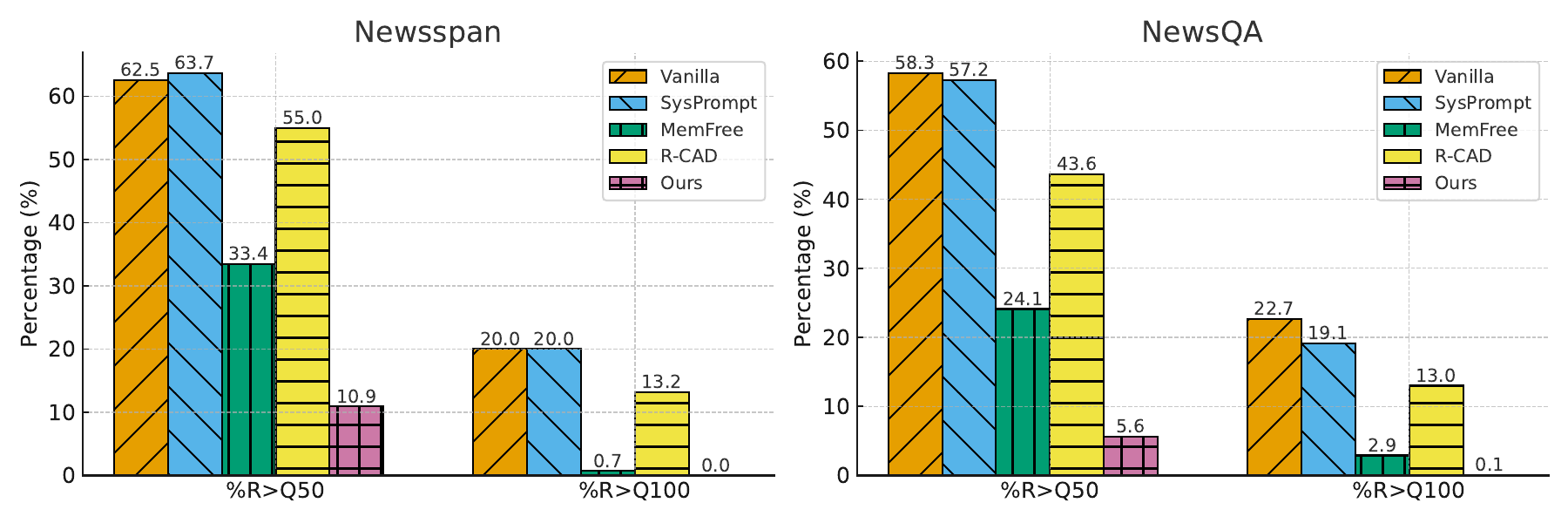}
    \caption{\method{} drastically outperforms other methods on long quote reduction.}
    \label{fig:longquote_perc}
\end{figure*}

\paragraph{(B) Dynamic rewriting with quote guidance}
We now detail the dynamic rewriting process of \method{} to ``scrub'' high-risk quotes from generated texts, overviewed in Alg.~\ref{alg:method}. Given the initial response $y \sim P_\text{gen}(\cdot | x)$ produced by the generation model $P_\text{gen}$ on prompt $x$, \method{} alternate between (A) quote extraction step and (B) rewriting step. 

We first extract verbatim quotes $q_1,\dots, q_n \leftarrow \mathcal{E}_C(y)$. If a quote longer than a pre-defined length threshold $\tau$ appears in $y$, the guided rewrite process is invoked. To conduct guided rewriting, we first create the rewrite instruction prompt $p_\text{rewrite}$ by feeding verbatim quotes into a pre-defined prompt template $p_r \leftarrow T(q_1,\dots, q_n)$ (detailed in \Sref{appsec:method_details}). Next, the rewrite model is instructed with this dynamic prompt to produce the rewritten output $y \sim P_\text{rewrite}(\cdot | p_r, y)$. Finally, we conduct the rewriting in an iterative manner: we extract quotes and proceed to rewriting repeatedly until long quote does not exist or a max iteration has been achieved.

The guided iterative rewriting process based on extracted quotes has several advantages. As we find in the ablation study (\Sref{sec:ablation}), quote guidance is crucial for reducing long quotes in rewritten outputs. Moreover, it is adaptive to varying levels of risk threshold by dynamically adjusting the number of rewrite iterations (\Sref{sec:scaling}). Finally, the rewrite model can scrub long quotes while retaining named entities that cannot be rewritten (\Sref{sec:long_quotes_analysis}), preserving utility. In contrast, MemFree decoding blocks all $n$-grams while keeping the already-generated $(n-1)$-gram prefix unchanged, risking utility while failing to remove the $(n-1)$-gram quote (\Sref{sec:failure_of_baselines}).

\paragraph{Certifying risk reduction through abstention}
If the max iteration for rewrite is achieved and rewrite model still fails to remove all long verbatim quotes, the \method{} system has the option of abstaining from producing a continuation. In this case, a refusal response will be used as the final generation $y$. In this case, our approach certifies that no quote from $C$ longer than $\tau$ will be generated.
This ensures that our \textit{soft removal} method obeys \textit{hard constraints}.

\input{tabs/mainres}
\begin{figure*}[ht]
    \centering
    \includegraphics[width=\linewidth,trim=0cm 0.5cm 0cm 0cm,clip=true]{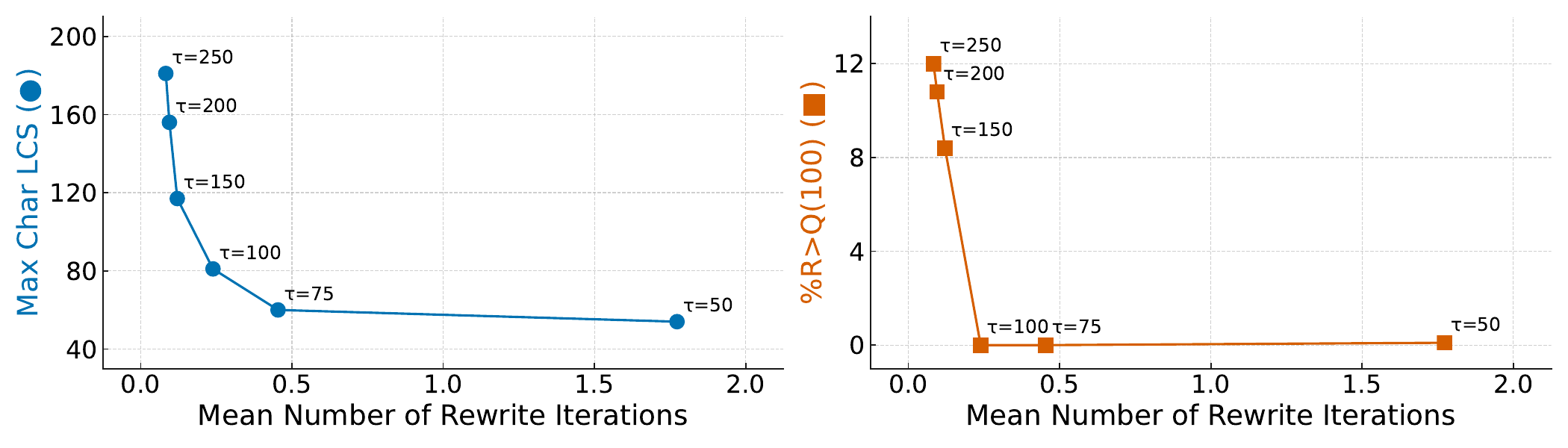}
    \caption{Inference-time adaptability of \method{} to different risk threshold $\tau$. As the risk threshold decreases, the mean number of rewrite iterations increases, and \method{} continues to reduce max character LCS and percentage of examples with quotes longer than 100 characters. 
    }
    \label{fig:scaling}
\end{figure*}

\section{Experimental Setup}
\label{sec:exp_setup}

\subsection{Task setup}
We expand from the task construction in the \textsc{CoTaEval} framework~\citep{wei2024evaluatingcopyrighttakedownmethods} to measure copyright infringement risk, information quality, and utility. For each document in the copyrighted corpus $C$, we use the first 200 tokens as the prompt to the model being evaluated and the next 200 tokens as the ground truth continuation. The goal of \textit{certified copyright takedown} is fourfold: generating responses that (1) does not contain any verbatim quote from the corpus (low worse-case infringement), (2) is not substantially similar to the grouth truth continuation (low average-case infringement), while (3) retaining the information in ground truth (high information quality), and (4) preserving factual knowledge in copyrighted corpus (high utility).

\subsection{Evaluation metrics}

\paragraph{Corpus-level infringment metric}
We propose a novel metric to quantify the \textbf{worst-case} infringement risk for a given model $M$ over a \textbf{large-scale} text corpus $C$: given a set of responses $\{y_i\}_{i=1}^N$ from $M$, $\%R>Q(\tau)$ measures the percentage of the responses that contain any quote of length greater than $\tau$:
\begin{equation*}
    \%R>Q(\tau) = \frac1N\sum_{i=1}^N\mathbbm{1}_{\{ s| s\subseteq y_i, s\in C, |s| >\tau \}\neq \emptyset},
\end{equation*}

\noindent where
$\mathbbm{1}_{\{\cdot\}}$ is the indicator function and 
$\subseteq$ denotes substring. This measures the empirical rate at which long quotes are generated, where a lower rate is more desirable. 

Unlike reference-based metrics such as longest common subsequence or ROUGE~\citep{lin2004rouge}, which only compare generated text to a specific reference, $\%R>Q(\tau)$ operates at the corpus level and consider long quotes from anywhere in $C$. This ensures a more comprehensive assessment of regurgitation risks in the \textbf{worst-case} over the entire corpus.

To efficiently compute this metric, we employ a Bloom filter of width $\tau$ and control the false positive rate to be lower than 0.001. In our experiments, we set $\tau$ to 50 or 100 characters as a strict bound.\footnote{Copilot's filter is reported to block verbatim matches longer than 150 characters~\cite{ippolito2022preventing}.} Importantly, this Bloom filter for metric calculation is different from the one used in \method{} quote extraction: for metric calculation, we use BF with width $\tau=50,100$; for \method, we use BF with width $n=25$.

\paragraph{Reference-based infringement metrics}

To comprehensively evaluate infringement, we also employ reference-based metrics against ground truth, including the maximum character-level and word-level longest common subsequence (\textit{\textbf{LCS}}), and word-level accumulated common subsequences (\textit{\textbf{ACS}}) across all test examples: given a set of response and corresponding ground truth $\{y_i, g_i\}_{i=1}^N$,
\begin{equation*}
    \text{Max}f = \max_i f(y_i, g_i),
\end{equation*}
\noindent where $f=\text{LCS}_\text{char}, \text{LCS}_\text{word},$ or ACS. We focus on the maximum LCS and ACS because they allow us to evaluate the \textbf{worse-case} outcome over all ground truth continuations. Finally, we also report the \textbf{\textit{win rate}} across 8 \textsc{CoTaEval}  metrics (detailed in \Sref{appsec:exp_details}---the probability that a given approach outpuerforms another approach on a random (metric, example) pair (see \citet{wei2024evaluatingcopyrighttakedownmethods} for details)---as an auxiliary measure for the \textbf{average-case} outcome of copyright takedown.

\paragraph{Information quality metrics} To evaluate the information quality of model predicted responses, we employ LLM-based evaluation of three aspects on a 5-point scoring scale: \textbf{\textit{Relevance}}, which whether the predicted continuation stays on-topic and appropriately responds to the given prompt; \textit{\textbf{faithfulness}}, assessing whether the predicted continuation contains information found in the ground truth; \textbf{\textit{hallucination}}, which identifies whether the predicted continuation includes any incorrect or fabricated information not present in the ground truth. The full details for evaluation is deferred to \Sref{appsec:info_eval}.

\paragraph{Utility metrics} Finally, to measure utility, i.e., whether the model still retains factual knowledge after mitigation, we follow \textsc{CoTaEval} and ask model questions related to the factual information in the copyrighted documents, and measure QA performance using the word-level \textbf{\textit{F1 score}} between predicted and ground truth answers.

\paragraph{Datasets}
We utilize 28K New York Times articles from the NewsSpan dataset~\citep{cheng2024dated} and 10K CNN-DailyMail articles from the NewsQA dataset~\citep{trischler2016newsqa} as two corpora of copyrighted content. For utility evaluation, we generate QA pairs for NewsSpan articles with GPT-4o (detailed in \Sref{appsec:newsspan_qa_gen}) and use NewQA QA pairs off-the-shelf. 

\paragraph{Models} In each experiment, we fine-tune Llama-3.1-8B-Instruct~\citep{dubey2024llama3herdmodels} on the target dataset as the generator model. We use the off-the-shelf Llama-3.1-8B-Instruct as the rewrite model. We use greedy decoding for all experiments.

\subsection{Baselines}
We compare our method with popular inference-time copyright takedown methods: the DBRX system prompt~\citep{DBRX2024}, MemFree decoding~\cite{ippolito2019comparison}, and Reverse Context Aware Decoding~\citep[R-CAD;][]{wei2024evaluatingcopyrighttakedownmethods}.

\paragraph{DBRX system prompt} System prompts are a set of instructions given to the LLMs before any user query. They have been used in production models to prevent the generation of certain types of content, e.g., copyrighted ones \citep{anthropic_system_prompts, DBRX2024}. The DBRX system prompt is shown to be the most effective variant of commonly used system prompts in \citet{wei2024evaluatingcopyrighttakedownmethods}.

\paragraph{MemFree Decoding} MemFree Decoding modify the decoding process to achieve \textit{hard removal} of all $n$-grams found in a given corpus $C$. At each step of decoding, it checks whether the next selected token would create an $n$-gram in $C$. If so, this token is blocked from being generated and the algorithm resamples the token with next-highest probability until no $n$-gram from $C$ will be created.

\paragraph{R-CAD} Context Aware Decoding~\citep{shi2023trusting} modifies the next token distribution and upweights the retrieved context of LLMs to reduce hallucination. R-CAD applies CAD in reverse to steer output away from retrieved context that may contain copyrighted material. When a prompt ${x}$ is fed into the model, R-CAD first retrieve a passage $c$ from $C$. It calculates the logits of generating the next token $y_i$ with $c$ presented in context, $\mathrm{logit}(y_i|{c},x, {y_{<i}})$. Instead of sample from the original logits $\mathrm{logit}(y_i|{x},{y_{<i}})$, it sample next token $y_i$ from the interpolated logits $\mathrm{softmax}[(1+\alpha)\mathrm{logit}(y_i|{x},{y_{<i}})-\alpha \mathrm{logit}(y_i|c,x,{y_{<i}})$ where the context $c$ is downweighted and $\alpha$ is the hyperparameter for the downweight strength.

We only consider inference-time methods because (1) our paper focus on inference-time methods, which are complementary training time methods, and (2) unlearning methods are shown to suffer great utility loss~\citep{wei2024evaluatingcopyrighttakedownmethods}. We defer further details and hyperparameters of \method{} and baselines to \Sref{appsec:method_details}.

\input{tabs/abstention}
\input{tabs/ablation}

\section{Experimental Results}

We now provide empirical evidence on the effectiveness of \method. We show that \method{} is both effective at worse-case copyright risk reduction and preserves utility, it is adaptable to varying levels of risk threshold at inference time, it can achieve certified risk reduction through abstention, and finally, the effectiveness of guided rewriting through an ablation study.
\label{sec:exp}
\subsection{Main results}
\label{sec:main_exp}

\paragraph{Infringement reduction and utility preservation} Shown in Fig.~\ref{fig:longquote_perc}, \method{} produce the least amount of long verbatim quotes on both datasets. Specifically, our method almost completely eliminates quotes longer than 100, compared to the vanilla decoded output with around 20\% long quotes. Table~\ref{tab:metric_results} corroborates this effectiveness of worst-case infringement reduction as \method{} achieves the lowest max LCS and ACS metrics across all settings. In the average case, our method is also comparable with baselines and is the top 2 methods in terms of win rate. We hypothesize that the average-case win rate is more effective on NewsSpan due to its larger size---and thus a richer set of extracted quotes from the Bloom filter. This suggests that \method{} is more effective when operating with practical, large-scale corpora. All methods except for R-CAD preserves information quality, and our method induce almost no utility loss in terms of the QA F1 score, demonstrating \method's potency in both infringement reduction and utility preservation.

\label{sec:scaling}

\begin{figure}[ht]
    \centering
    \includegraphics[width=\linewidth,trim=0cm 0.52cm 0cm 0.4cm,clip=true]{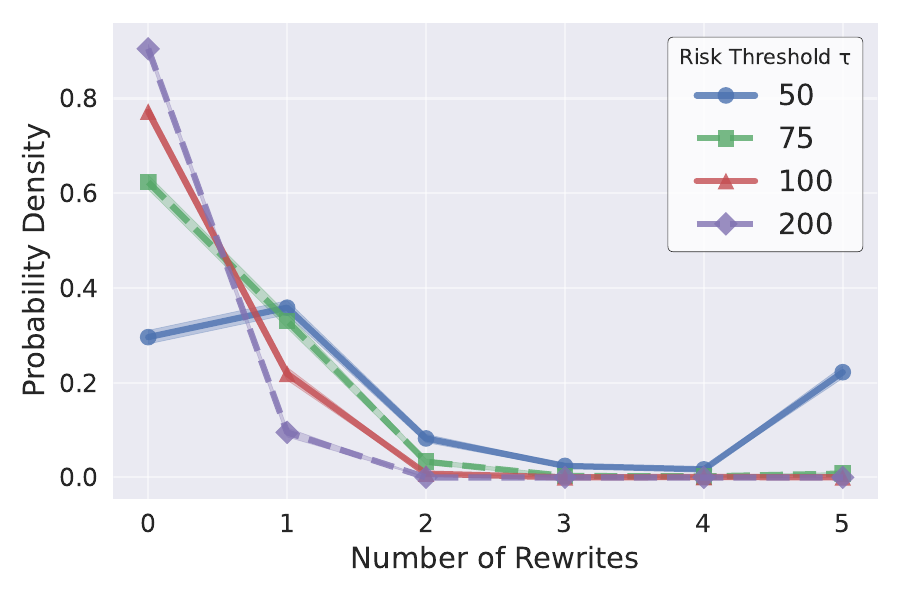}
    \caption{Distribution of number of rewrites under different risk threshold $\tau$. Given a smaller (thus more stringent) $\tau$, the distribution of rewrite shifts to the right.}
    \label{fig:num_rewrite}
\end{figure}

\paragraph{Inference-time adaptability} To demonstrate the inference-time adaptability of \method, we run our method on NewsSpan while varying the risk threshold $\tau$. Shown in Fig.~\ref{fig:scaling} and Fig.~\ref{fig:num_rewrite}, as $\tau$ decreases, our method continually improves both max LCS and $\%R>Q(100)$ metrics at the cost of increased number of rewrite iterations. Interestingly, as the threshold decreases to 100, $\%R>Q(100)$ quickly drops to a near-zero value, indicating the effectiveness of long quote reduction.

\subsection{Certified risk reduction through abstention}
\label{sec:abstention}
Table~\ref{tab:abstain} demonstrates \method{} can achieve certified risk reduction through the incorporation of the abstention mechanism. At $\tau=50$, we achieve perfect scores of 0.0\% on the $\%R>Q$ metrics. Abstention also have a positive effect on the Max LCS metric, pushing it down to below 50. Because \method{} already performs well on $\%R>Q$ without abstention, incorporating abstention only imposes a small cost on information quality, reducing the relevance and faithfulness scores. On the other hand, abstention leads to slightly better hallucination scores since abstained responses do not hallucinate.

\subsection{Ablations of the guided rewrite objective}
\label{sec:ablation}
To verify the effectiveness of the quote-guided rewriting approach, we conduct ablation by conducting the rewrite process without quote guidance. Shown in Table~\ref{tab:ablate}, the ablated method lead to both a higher rate of $\%R>Q(50)$ and a higher maximum char LCS metric across two datasets, indicating the value of guiding the ``scrubbing'' process with explicit high-risk quotes.

\section{Analysis}

\subsection{The Remaining Long Quotes}
\label{sec:long_quotes_analysis}
Eliminating all verbatim quotes from copyrighted sources longer than a threshold $\tau$, while effective at reducing copyright risks, may lead to overprotection.
It is likely reasonable to preserve certain types of long quotes, e.g., named entities or phrases that are crucial for conveying the information in the copyrighted source. As an example, ``the Fundamentalist Church of Jesus Christ of Latter-day Saints'' is a named entity spanning 62 characters that appeared in NewsQA. Since \method{} without abstention measures a small but non-zero rate of $\%R>Q(50)$, we conduct analysis to answer this question: how many remaining quotes of \method{} contain named entities that are difficult to rewrite?

\begin{figure}
    \centering
    \includegraphics[width=\linewidth]{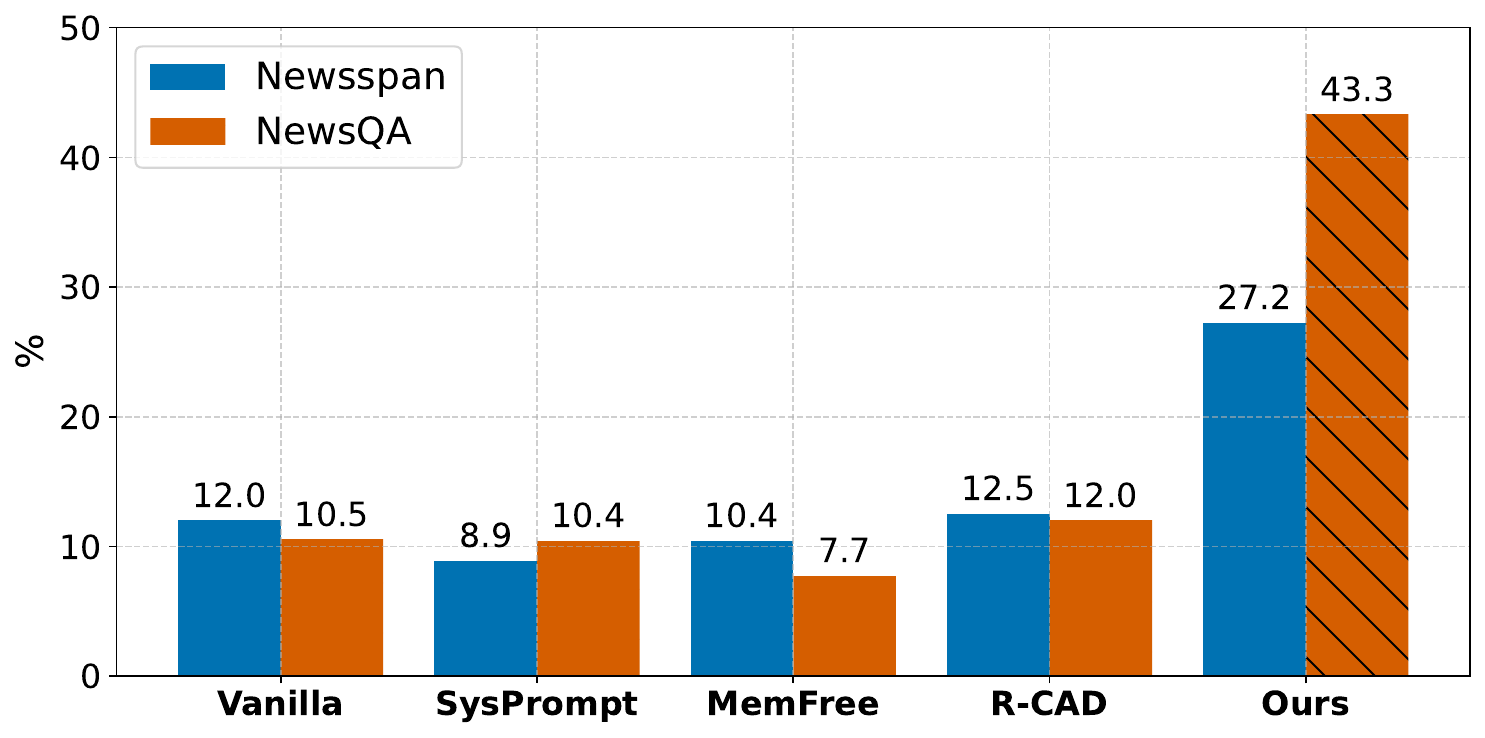}
    \caption{Percentage of long quotes ($\geq$50 characters) that contain a long named entity ($\geq$30 characters). A high rate of long named entity indicates that a notable portion of remaining quotes are difficult to rewrite, thus most quotes that \textit{can} be rewritten \textit{have} been rewritten. 
    }
    \label{fig:named-ent}
\end{figure}

Shown in Fig.~\ref{fig:named-ent}, we find that the remaining long quotes ($\geq$50 characters) after running \method{} contain a significantly higher percentage of long named entities ($\geq$30 characters, determined by spaCy~\citep{spacy2}) compared to vanilla decoding and other baselines. This indicates that most long quotes that \textit{can} be rewritten have been rewritten by \method, and thus a larger portion of the remaining quotes contain named entities. We find that the quote-guided rewriting instruction of \method{} behaves like a ``soft constraint'' and the rewrite model has the option to retain quotes that are difficult to rewrite, which is advantageous for utility preservation. We provide qualitative examples of long quotes in \Sref{appsec:long_quote_ex}.

\subsection{Failure Modes of R-CAD and MemFree decoding}
\label{sec:failure_of_baselines}

Because R-CAD and MemFree decoding modifies the output distribution directly, they are at risk for degenerated response quality. For example, we find that R-CAD sometimes generate texts with missing spaces or nonexistent words:

\begin{tcolorbox}[colframe=black!50, colback=gray!10, boxsep=1pt, left=2pt, right=2pt, top=2pt, bottom=2pt]
Maximum sustained windsstrengthened some during the day to145 mph (233 kph).
\end{tcolorbox}

\begin{tcolorbox}[colframe=black!50, colback=gray!10, boxsep=1pt, left=2pt, right=2pt, top=2pt, bottom=2pt]
...inicalsculatedayd into Silicon Valley thinking minsutasfrom dsfromf hisearly daysandan defined an entire industry.
\end{tcolorbox}

\noindent Moreover, as reported in \citet{wei2024evaluatingcopyrighttakedownmethods}, R-CAD is at risk at significant utility loss when the ground truth document is retrieved, further exacerbating the utility risk for R-CAD.

On the other hand, MemFree decoding suffers from similar token-perturbation issues since certain tokens are blocked from being generated:

\begin{tcolorbox}[colframe=black!50, colback=gray!10, boxsep=1pt, left=2pt, right=2pt, top=2pt, bottom=2pt]
Bill is forecast\quad to approach Bermuda late Friday night or Saturday.
\end{tcolorbox}

\begin{figure}
    \centering
    \includegraphics[width=\linewidth,trim=0cm 0.3cm 0cm 0cm,clip=true]{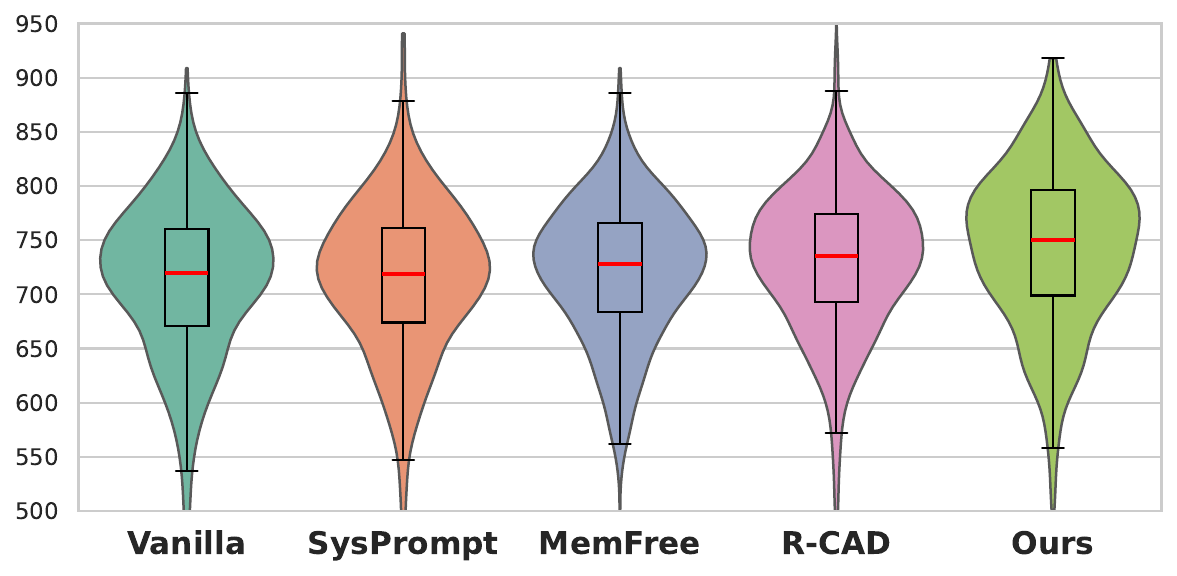} 
    \caption{Levenshtein distance between ground truth and predicted responses of different prevention methods. MemFree decoding only marginally increase the Levenshtein distance, while R-CAD and \method{} are more effective at preventing near-verbatim matches with the copyrighted source.}
    \label{fig:lev_analysis}
\end{figure}

\noindent In this sentence, an `ed' is missing after `forecast', and there is an extra space. This not only creates fluency issue but also still induce infringement risk because most of the text is unchanged, as shown by the smaller increase of Levenshtein distance from vanilla, compared to R-CAD and \method{} (Fig.~\ref{fig:lev_analysis}). Our method does not suffer from these issues as we do not manipulate local token distributions.

Interestingly, while \method's rewrite process rely only on verbatim quotes that need to be removed, it does not suffer the same issue of limited Levenshtein distance that MemFree decoding have. We surmise two factors contributes to this advatageous behavior: (1) the dynamic LLM-based rewriting process allow a form of \textit{global} planning, where the entire text, instead of just a few tokens, is reproduced, and (2) the fixed-width Bloom filter design (\Sref{subsec:our:approach}) enables near-verbatim ``stitched quotes'' to be extracted, expanding the candidate set for rewrite.  

\section{Discussion and Future Work}

In \Sref{sec:exp}, we provide rich empirical evidence that our \method{} method enables models to use knowledge while ensuring that responses are transformative, disallowing generations that are excessively copied and therefore effectively reducing copyright infringement risk. 
Our approach is flexible, with a dynamic number of rewrites and adjustable risk thresholds, but can still enforce hard limits through abstentions, achieving \textit{certified} copyright takedown.
Our method can also easily accommodate changing corpora (e.g. resulting from new licensing agreements) and effective at a large scale.

Our proposed metrics focus on evaluating the \textit{worst-case} infringement outcome over massive corpora, an important but overlooked aspect for reducing copyright risk for LLMs---even if the average behavior of a model appears acceptable, a single instance of significant infringement could trigger costly legal challenges or severe reputational damage. We thus underscore the need for further research to better understand and mitigate worst-case infringement risks in large language models.

Our work focuses on developing a certified approach to eliminate \textit{verbatim} regurgitation while preserving quality and utility—an essential step toward aligning model outputs with the \textit{transformativeness} principle of fair use. However, we emphasize that this is a crucial first step but likely insufficient measure for fully mitigating infringement risks. Beyond verbatim copying, non-literal reproduction~\citep{chen2024copybenchmeasuringliteralnonliteral} poses additional challenges, where achieving certified risk reduction remains an open problem.

Our goal of removing \textit{all} long verbatim quotes might lead to overprotection in some cases, e.g., removal of long named entities that is permitted. However, we empirically show that even after \method{} eliminates long quotes, it only entails minimal utility loss (\Sref{sec:exp}), suggesting that the overprotection problem is minor in practice.

Finally, as a plug-and-play, inference-time solution, \method{} seamlessly integrates with existing LLMs and are complementary to training-time mitigation approaches. Future work could explore the synergy between training- and inference-time methods to develop more comprehensive copyright-compliant LLM frameworks.

\section*{Limitations}

While \method{} effectively reduces verbatim regurgitation, eliminating direct quotations alone is a necessary but not sufficient condition for mitigating copyright risk. Non-literal copying~\citep{chen2024copybenchmeasuringliteralnonliteral}, such as paraphrased or stylistically similar outputs, remains an open challenge and requires further collaborative investigation between the AI and legal communities. Additionally, while we employ a Bloom filter for efficient quote detection, this component can be replaced with alternative data structures, such as suffix arrays (e.g., Infini-gram~\citep{liu2024infinigram}), which we have not explored. Lastly, while we conduct analysis on overprotection and unrewritable quotes consists of named entities, further analysis and deliberations can be done to mitigate the overprotection problem at a finer granularity. For example, instead of using a fixed risk threshold $\tau$ for all responses, $\tau$ can be determined dynamically based on the context.

\section*{Ethical Considerations}
Our work aims to mitigate copyright risks by preventing verbatim regurgitation while preserving text utility, aligning with the principle of transformativeness in fair use. However, defining copyright boundaries in generative models remains complex, especially regarding non-literal reproduction. As automated copyright mitigation becomes more widespread, it is essential to gauge the robustness of these methods, identify failure modes, and balance the rights of creators, LLM developers, and the public to foster more responsible and equitable deployment of AI systems.

\section*{Acknowledgments}
This work was supported in part by a grant from Amazon through the AI2AI program. 
We thank Yining Lu and the JHU CLSP community for their helpful feedback. 


\bibliography{bib/custom, bib/ref}

\newpage

\appendix

\onecolumn

\begin{center}
{\Large \textbf{Supplemental Material}}
\end{center}

\section{Experiment details}
\label{appsec:exp_details}

At the beginning of our experiment, we finetune Llama3.1-8B-Instruct using LlamaFactory~\citep{llamafactory} under the pertaining mode with a learning rate of 1.0e-5 for 3 epochs. We set all the parameters to trainable. The NewsQA dataset is used under the MIT License.

\paragraph{Metrics for average win rate calculation} We use the following 8 metrics from \citet{wei2024evaluatingcopyrighttakedownmethods} to calculate the average win rate between copyright takedown methods:
\begin{itemize}
    \item ROUGE-1 score
    \item ROUGE-L score
    \item Semantic similarity
    \item Character-level LCS
    \item Word-level LCS
    \item Levenshtein distance
    \item Minhash similarity
\end{itemize}
We refer readers to \citet{wei2024evaluatingcopyrighttakedownmethods} for further details.

\section{Method details}
\label{appsec:method_details}
\subsection{System prompt}
We use the DBRX system prompt~\citep{DBRX2024} for the system prompt method:

\begin{tcolorbox}[colframe=black!50, colback=gray!10, boxsep=1pt, left=2pt, right=2pt, top=2pt, bottom=2pt]
You are a helpful, respectful and honest assistant. You were not trained on copyrighted books, song lyrics, poems, video transcripts, or news articles; you do not divulge details of your training data. You do not provide song lyrics, poems, or news articles and instead refer the user to find them online or in a store.
\end{tcolorbox}

\subsection{\method{} details}

We use Bloom filters of width 25 for quote extraction, and set the max number of rewrite iterations to 5. We provide the prompt template for rewrite instruction below:

\begin{tcolorbox}[colframe=black!50, colback=gray!10, boxsep=1pt, left=2pt, right=2pt, top=2pt, bottom=2pt]
[Insert the text to paraphrase here] Paraphrase the provided text while preserving its meaning, using different words and sentence structures. Ensure clarity, coherence, and maintain any specified tone or style. Importantly, completely rewrite this part of text: [Insert the longest quoted segment here]
\end{tcolorbox}

\subsection{Hyperparameter selection for MemFree decoding and R-CAD}

Following~\citet{wei2024evaluatingcopyrighttakedownmethods}, for MemFree decoding we set $n$, the width of token n-grams, to 6. For R-CAD, we set $\alpha$, the weight of adjustment, to $1.0$. Decreasing $n$ or increasing $\alpha$ can lead to better results in infringement evaluation, but will result in decrease in utility scores and quality of generated text. We choose these values so that their utility scores stay at a reasonable level and the information quality of the generated text does not diverge too much. For example, when increasing $\alpha$ from 1 to 3 for R-CAD, relevance score drops from 2.6 to 2.1, faithfulness from 1.8 to 1.5, and hallucination from 2.0 to 1.7.

\section{Newsspan question generation}
\label{appsec:newsspan_qa_gen}
To conduct question-answer pairs generation for NYT articles sourced from the NewsSpan dataset~\citep{cheng2024dated}, we use the following prompt on GPT-4o~\citep{gpt4}: 
\begin{tcolorbox}[colframe=black!50, colback=gray!10, boxsep=1pt, left=2pt, right=2pt, top=2pt, bottom=2pt]
\begin{lstlisting}
"messages":[
    {"role": "system", "content": "**Task**:\n- Write a factual, non-ambiguous question based on the article. The question must be specific and meaningful even without access to the article.\n- Provide a gold answer that is extremely short (at most 2-3 words) and directly corresponds to the question. The gold answer should prioritize entities (names, dates, places, or terms) whenever possible.\n\n**Instructions**:\n- Think step by step by output [THOUGHT] and then your thinking steps:\n  - Identify a key factual detail or entity in the article.\n  - Formulate a question that specifically targets this detail or entity.\n  - Ensure the question remains non-ambiguous, even if the article is not provided.\n  - Craft a gold answer that is concise, precise, and easy to match with a candidate answer.- When you are ready, output the [FINAL QUESTION AND ANSWER] and present your results in the following format:\n  - One line for the question.\n  - One line for the gold answer."},
    {"role": "user", "content": "{article}"}
],
\end{lstlisting}
\end{tcolorbox}
The above prompts ensures the gold answer that are generated are short and easy to match. Here are some examples of the generated QA pairs:

\begin{tcolorbox}[colframe=black!50, colback=gray!10, boxsep=1pt, left=2pt, right=2pt, top=2pt, bottom=2pt]
\begin{lstlisting}
1.	Which court denied Adnan Syed a new trial?
Maryland's highest court
2.	What is the NBA trade deadline date mentioned in the article?
Feb. 7
3.	Which company received conditional approval from Chinese regulators for a merger with SABMiller?
Anheuser-Busch
4.	On what date was the Kentucky Derby rescheduled in 2020 due to the coronavirus pandemic?
Sept. 5
5.	On what date did the judge sign the search warrant related to Hillary Clinton's email investigation?
Oct. 30
\end{lstlisting}
\end{tcolorbox}

\clearpage
\section{Information Quality Evaluation details}
\label{appsec:info_eval}
To evaluate the information quality of model generated continuations, we employ a LLM-based evaluation scheme and prompt Llama3.3-70B-Instruct with relevant evaluation instructions to produce a score on a 5-point scale. Specifically, we adapt judge prompts from \citet{zheng2023judging} and use the following prompt for evaluation.

Relevance evaluation:
\begin{tcolorbox}[colframe=black!50, colback=gray!10, boxsep=1pt, left=2pt, right=2pt, top=2pt, bottom=2pt]
\begin{lstlisting}
**Task**:  \nYou will be provided with:\n- **Prompt**: The initial input prompt.\n- **Ground Truth Continuation**: The expected response continuation.\n- **Predicted Continuation**: The generated response.\n\n**Evaluation Criteria for Relevance**:  \nAssess whether the predicted continuation stays on-topic and appropriately responds to the given prompt. Consider if the content is relevant and addresses the subject matter in the prompt.\n\n**Scoring**:  \nAssign a score from **1 to 5**, where:\n- **1 - Poor**: The predicted continuation is mostly off-topic and fails to address the prompt.\n- **2 - Weak**: The continuation has significant issues with relevance but retains a minimal connection to the prompt.\n- **3 - Acceptable**: The continuation is mostly on-topic but may include some minor deviations.\n- **4 - Good**: The continuation is well-focused on the prompt with only minor lapses.\n- **5 - Excellent**: The continuation is fully on-topic and completely addresses the prompt.\n\n**Output Format**:  \nBegin your evaluation by providing a short explanation. Be as objective as possible. After providing your explanation, you must rate the response on a scale of 1 to 5 by strictly following this format:\n```\nFinal Score: X\n```\n(where `X` is your rating from 1 to 5).
\end{lstlisting}
\end{tcolorbox}

\vfill
\break

Faithfulness evaluation:
\begin{tcolorbox}[colframe=black!50, colback=gray!10, boxsep=1pt, left=2pt, right=2pt, top=2pt, bottom=2pt]
\begin{lstlisting}
**Task**:  \nYou will be provided with:\n- **Prompt**: The initial input prompt.\n- **Ground Truth Continuation**: The expected response continuation.\n- **Predicted Continuation**: The generated response.\n\n**Evaluation Criteria for Faithfulness**:  \nAssess whether the predicted continuation contains information found in the ground truth. Ensure that it accurately reflects the content and avoids introducing unsupported claims or deviations from the ground truth.\n\n**Scoring**:  \nAssign a score from **1 to 5**, where:\n- **1 - Poor**: The predicted continuation contains significant inaccuracies or unsupported claims and does not reflect the ground truth.\n- **2 - Weak**: The continuation has notable issues with accuracy and alignment with the ground truth.\n- **3 - Acceptable**: The continuation is mostly accurate with minor inaccuracies or omissions.\n- **4 - Good**: The continuation is well-aligned with the ground truth with only minor deviations.\n- **5 - Excellent**: The continuation is fully faithful to the ground truth with no inaccuracies.\n\n**Output Format**:  \nBegin your evaluation by providing a short explanation. Be as objective as possible. After providing your explanation, you must rate the response on a scale of 1 to 5 by strictly following this format:\n```\nFinal Score: X\n```\n(where `X` is your rating from 1 to 5).
\end{lstlisting}
\end{tcolorbox}

\vfill
\break

Hallucination evaluation:
\begin{tcolorbox}[colframe=black!50, colback=gray!10, boxsep=1pt, left=2pt, right=2pt, top=2pt, bottom=2pt]
\begin{lstlisting}
**Task**:  \nYou will be provided with:\n- **Prompt**: The initial input prompt.\n- **Ground Truth Continuation**: The expected response continuation.\n- **Predicted Continuation**: The generated response.\n\n**Evaluation Criteria for Hallucination**:  \nIdentify whether the predicted continuation includes any incorrect or fabricated information not present in the ground truth. Evaluate if there are any hallucinated details that misrepresent or add unwarranted content.\n\n**Scoring**:  \nAssign a score from **1 to 5**, where:\n- **1 - Poor**: The predicted continuation contains numerous hallucinations and fabricated details not supported by the ground truth.\n- **2 - Weak**: The continuation includes several instances of hallucination, significantly affecting its credibility.\n- **3 - Acceptable**: The continuation has minor hallucinated elements, but these do not majorly undermine the content.\n- **4 - Good**: The continuation contains minimal hallucinations with mostly accurate representation.\n- **5 - Excellent**: The continuation is free of hallucinations and completely aligns with the ground truth.\n\n**Output Format**:  \nBegin your evaluation by providing a short explanation. Be as objective as possible. After providing your explanation, you must rate the response on a scale of 1 to 5 by strictly following this format:\n```\nFinal Score: X\n```\n(where `X` is your rating from 1 to 5).
\end{lstlisting}
\end{tcolorbox}

\section{Qualitative examples of long quotes after rewriting}
\label{appsec:long_quote_ex}

We show qualitative examples of long quotes that are still present in the model generation below. Many of these long quotes contain long named entities that are difficult to rewrite, but are also likely low risk for copyright infringement.

NewsSpan:
\begin{tcolorbox}[colframe=black!50, colback=gray!10, boxsep=1pt, left=2pt, right=2pt, top=2pt, bottom=2pt]
\begin{lstlisting}
<quote1>Should healthy people be wearing masks when they're outside to protect themselves and others?
<quote2> for The Guardian, said he was "body slammed" by Greg Gianforte, a Republican candidate 
<quote3> of communication between the incoming administration and the Russian government.
<quote4>s. The Federal Reserve and the New York State Department of Financial Services 
<quote5>
...CBS News Magazine "60 Minutes" features the story of Beckett Brennan, a 
<quote6> Dr. Donald Hensrud, director of the Mayo Clinic's Healthy Living Program. 
<quote7> Chris Christie of New Jersey, who briefly led the Trump transition team, 
<quote8> Chris Christie of New Jersey, who briefly led the Trump transition team, 
<quote9> "If I Had a Hammer," "Goodnight Irene," and "Kisses Sweeter Than Wine," 
<quote10> a billion acres in the Arctic, Pacific, Atlantic, and Gulf of Mexico. T
\end{lstlisting}
\end{tcolorbox}

\vfill
\break

NewsQA:
\begin{tcolorbox}[colframe=black!50, colback=gray!10, boxsep=1pt, left=2pt, right=2pt, top=2pt, bottom=2pt]
\begin{lstlisting}
<quote1>s motivated by a person's actual or perceived gender, sexual orientation, gender identity, or disability. 
<quote2> the US Department of Health and Human Services and the Centers for Disease Control and Prevention, 
<quote3> David Petraeus, the top US commander in Iraq, and Ryan Crocker, the US ambassador to 
<quote4>s.
The FDA is warning consumers to immediately stop using 14 Hydroxycut products, 
<quote5> Rear Admiral Gregory Smith, the U.S. military's chief spokesman in Iraq, 
<quote6>to the Fundamentalist Church of Jesus Christ of Latter-day Saints (FLDS) 
<quote7> the Fundamentalist Church of Jesus Christ of Latter-day Saints (FLDS). 
<quote8> the Fundamentalist Church of Jesus Christ of Latter-day Saints (FLDS), 
<quote9>t:
The Fundamentalist Church of Jesus Christ of Latter-day Saints, a 
<quote10> Ralph Nicoletti, 18, Michael Contreras, 18, and Brian Carranza, 21, 
\end{lstlisting}
\end{tcolorbox}

\section{Examples}

We now provide an example of two rounds of iterative rewriting. Extracted quotes are highlighted.

First Rewrite:
\begin{tcolorbox}[colframe=black!50, colback=gray!10, boxsep=1pt, left=2pt, right=2pt, top=2pt, bottom=2pt]

"messages": [
\{
  "role": "user",
  "content": "chemical \hl{nickname, was part of the Anfal campaign, in which the Hussein regime killed a}n estimated 100,000 Iraqi Kurds with poison gas and went on to round up and execute thousands more. \hl{Al-Majeed was sentenced to death in }1988 for his role in the Anfal campaign, but the \hl{sentence was commuted to life in prison. After the U.S.-led invasion of Iraq in 2003,} he was arrested by U.S. forces and put on trial in Iraq. He was convicted in 2007 \hl{of crimes against humanity and genocide }and sentenced to death. \hl{He was hanged after having been convicted on 13 counts of killings and genocide}, \hl{an Iraqi government spokesman said. He was} \hl{sentenced to death in four separate trials, including one that focused on his involvement in a poison gas attack against Iraqi Kurds that killed about 5,000 people. His execution had been delayed for political rather than legal reasons. It is not clear what change, if any, led to the}
  
  Paraphrase the provided text while preserving its meaning, using different words and sentence structures. Ensure clarity, coherence, and maintain any specified tone or style."
\}
],

"rewritten text": "The notorious chemical nickname was a key figure in the Anfal campaign, a brutal operation carried out by \hl{Saddam Hussein's regime} \hl{that resulted in the deaths of a}pproximately 100,000 Iraqi Kurds through poison gas and the execution of thousands more. In 1988, Al-Majeed was condemn\hl{ed to death for his role in }the atrocities, but hi\hl{s sentence was commuted to life imprisonment.} Following \hl{the US-led invasion of Iraq in 2003,} he was apprehended by US forces and put on trial in Iraq. In 2007, Al-Majeed was found \hl{guilty of crimes against humanity and genocide,} and \hl{subsequently sentenced to }death. His \hl{execution was carried out} \hl{after he was convicted o}n 13 counts of murder and \hl{genocide, according to an I}raqi \hl{government spokesperson}. \hl{Al-Majeed was sentenced to death in four separate trials}, \hl{including one that focused on his involvement in a poison gas attack} \hl{that claimed the lives of a}round 5,000 Iraqi Kurds. The delay in his execution was reportedly due to"
\end{tcolorbox}

Second Rewrite:
\begin{tcolorbox}[colframe=black!50, colback=gray!10, boxsep=1pt, left=2pt, right=2pt, top=2pt, bottom=2pt]

"messages": [
{
  "role": "user",
  "content": "The notorious chemical nickname was a key figure in the Anfal campaign, a brutal operation carried out by \hl{Saddam Hussein's regime} \hl{that resulted in the deaths of a}pproximately 100,000 Iraqi Kurds through poison gas and the execution of thousands more. In 1988, Al-Majeed was condemn\hl{ed to death for his role in }the atrocities, but hi\hl{s sentence was commuted to life imprisonment.} Following \hl{the US-led invasion of Iraq in 2003,} he was apprehended by US forces and put on trial in Iraq. In 2007, Al-Majeed was found \hl{guilty of crimes against humanity and genocide,} and \hl{subsequently sentenced to }death. His \hl{execution was carried out} \hl{after he was convicted o}n 13 counts of murder and \hl{genocide, according to an I}raqi \hl{government spokesperson}. \hl{Al-Majeed was sentenced to death in four separate trials}, \hl{including one that focused on his involvement in a poison gas attack} \hl{that claimed the lives of a}round 5,000 Iraqi Kurds. The delay in his execution was reportedly due to"
  
  Paraphrase the provided text while preserving its meaning, using different words and sentence structures. Ensure clarity, coherence, and maintain any specified tone or style. Importantly, completely rewrite this part of text: \hl{including one that focused on his involvement in a poison gas attack}"
}
],

"rewritten text": "Here's a paraphrased version of the text: The infamous chemical engineer was a key perpetrator of the Anfal campaign, a brutal operation orchestrated by \hl{Saddam Hussein's regime} \hl{that led to the deaths of a}pproximately 100,000 Iraqi Kurds through chemical warfare and the execution of thousands more. In 1988, \hl{he was sentenced to death for his role in }the atrocities, but hi\hl{s sentence was commuted to life imprisonment.} Following \hl{the US-led invasion of Iraq in 2003,} he was apprehended by US forces and put on trial in Iraq. In 2007, he was found \hl{guilty of crimes against humanity and genocide,} and \hl{subsequently sentenced to }death. His \hl{execution was carried out} \hl{after he was convicted o}n 13 counts of murder and \hl{genocide, according to an I}raqi \hl{government spokesperson.} The delay in his execution was reportedly due to various factors, \hl{including the complexity of the }trials and the need for thorough investigations."
\end{tcolorbox}

\newpage
\subsection{Comparison between \method{} and baselines}
We now provide an example generation of \method{} compared with baselines when responding to the same prompt.

\begin{table}[ht]
\footnotesize
\centering
\begin{tabular}{p{2cm} p{11cm} p{1.5cm}}
\toprule
Prompt & WASHINGTON (CNN)  -- Boys may soon be able to get Gardasil, the vaccine given to girls and young women to prevent infection by four types of human papillomavirus. Gardasil, a vaccine against human papillomavirus, would be given to boys exactly as it is to girls. A Food and Drug Administration advisory committee voted Wednesday to recommend that the vaccine be made available to boys and young men aged 9 to 26 for protection against genital warts caused by HPV. The vaccine protects against four types of HPV, and two of those are believed to be responsible for 70 percent of cervical and anal cancers, and HPV-associated penile and throat-and-neck cancers. The other two cause 90 percent of genital warts cases, researchers say. At Wednesday's advisory committee meeting, pharmaceutical giant Merck \& Co., maker of Gardasil, presented data from three clinical trials that the company claims supports broadening the distribution of the vaccine to include males. The trials included more than & Length of Longest Quoted Segment\\
\midrule
SysPrompt & 7,000 boys and men, according to Dr. Barbara Howe, a Merck researcher. The trials showed that the vaccine was 9\hl{0 percent effective in preventing }genital warts, an\hl{d there were no serious adverse events related to the vaccine,} she said. The vaccine would be given \hl{in boys and young men exactly as it }is \hl{given to girls and young women: three doses over a period of six months,} she said. The vaccine is not 100 percent effective, and there are risks associated with it, said Dr. Barbara Howe, a Merck researcher. The vaccine is not a cure for HPV, she said, and it does not treat existing HPV infections. It also does not protect agains\hl{t sexually transmitted diseases o}ther than genital warts, she said. There were som\hl{e side effects, including} \hl{pain at the injection site,} dizziness, nausea, fever, swelling of the lymph nodes in the arm where the shot was given, and muscle aches, she said. There were also some allergic reactions, & 72 \\
 
\midrule
MemFree & 5,400 boys and 23,000 girls and \hl{women. According to the d}ata, \hl{there were no serious adverse }reactions to the vaccine, and the most common side effects were \hl{pain at the injection site a}s well as dizziness, nausea, vomiting and fainting. The vaccine would be given in boys and young men i\hl{n three doses, over a period of six }weeks, exactly as it is \hl{given to girls and young wom}an. Merck spokesman Pam Eisele said that the cost of the vaccine will remain the same as for girls, \$130 per dose. Three doses are required, so the total cost to boys and young men would be \$390. The vaccine is not meant to be a cure-all for \hl{sexually transmitted infections. ""I}t's not a license to be promiscuous,"" said Dr. Anne Marie Lennon, a gynecologic \hl{oncologist at the University of Pennsylvania's }Abramson Cancer Center. ""It's not a get-out-of-jail-free card. You still have to practice safe & 47 \\
\midrule
R-CAD & 7,200 boys and men, half of whom were given the vaccine and half of whom were given a placebo. According to Merck, the data showed that the vaccine was 9\hl{0 percent effective in preventing }genital warts in both males and females one to two years after vaccination, and that it maintaine\hl{d more than 99 percent of }its effectiveness in preventing disease caused by the virus in the female partners of male recipients. However, the data also showed that the vaccine was no\hl{t effective in preventing disease }in male recipients' partners if the partners themselves were infected with HPV at the time of sexual contact. The data also showed that the vaccine did not prevent genital warts in men, although it did preven\hl{t genital warts caused by HPV }in men \hl{who had not previously been i}nfected. Critics of the vaccin\hl{e have raised concerns that }broadening its distribution will overburden an already-strained health care system with unnecessary vaccinations and office visits, and that the vaccine does not protect against the most dangerous forms of HPV. The & 34 \\
\midrule
\method{} & According to Dr. Barbara Howe, a researcher at Merck, \hl{a substantial number of }male participants, approximately 5,400, have been involved in vaccine-related studies. These studies have yielded remarkable results, with 70\% of vaccinated men experiencing a significa\hl{nt reduction in genital warts,} resulting in an 89\% decrease. Moreover, the vaccine has demonstrated exceptional efficacy in reducing the incidence of genital warts-causing HPV infection, with a notable 90\% reduction observed in vaccinated men. The vaccine is administered in a three-dose regimen, with the CDC estimating that each dose will cost around \$130. It's worth noting that the vaccine does not offer complete protection against all types of HPV, necessitating the continued use of Pap tests to screen for cervical cancer. Interestingly, the vaccine has been available \hl{to girls and young women }since June 2006, and its benefit\hl{s have been well-documented.} However, the data on male participants is a more recent development. & 30 \\
\bottomrule
\end{tabular}
\end{table}

\end{document}

%% file: alg.tex
\begin{algorithm}[ht]
  \small
  \caption{\method 
    \label{alg:method}}
  \begin{algorithmic}[1]
    \Require{prompt $x$, generation model $P_\text{gen}$, rewrite model $P_\text{rewrite}$, quote extractor $\mathcal{E}_C$, prompt template $T$}
    \Ensure{threshold $\tau$, max iteration $i_\text{max}$}

    \State {$y \sim P_\text{gen}(\cdot | x)$} \Mycomment{The initial response}
    
    \Let {$i$}{$0$}
    \While {$i\leq i_\text{max}$}
        \Let{$q_1,\dots, q_n $}{$\mathcal{E}_C(y)$} \Mycomment{Identify verbatim quotes}
        \If{$\text{maxlen}(q_1\dots q_n) < \tau$}
            break 
        \EndIf
        \Let {$p_r$} {$T(q_1,\dots, q_n)$} \Mycomment{Form scrubbing prompt} 
        
        \State $y \sim P_\text{rewrite}(\cdot | p_r, y)$ \Mycomment{Scrub the verbatim quotes}
        \State $i++$
    \EndWhile

    \If{$\text{maxlen}(q_1\dots q_n) \geq \tau$}
        \Mycomment{Optional: abstention}
        \Let{$y$}{Sorry, I am unable to respond.}
    \EndIf

    \State \textbf{return} $y$
  \end{algorithmic}
\end{algorithm}

%% file: tabs/mainres.tex
\begin{table*}[ht]
\centering
\resizebox{\linewidth}{!}{
\begin{tabular}{@{}llcccccccc@{}}
\toprule
\multirow{2}{*}{Dataset} & \multirow{2}{*}{Method} & \multicolumn{4}{c}{\textit{\textbf{Infringement (against ground truth continuation)}}} & \multicolumn{3}{c}{\textit{\textbf{Info Quality}}$\uparrow$} & \multicolumn{1}{c}{\textit{\textbf{Utility}}$\uparrow$} \\
\cmidrule(l){3-6} \cmidrule(l){7-9} \cmidrule(l){10-10}  
& & Max LCS$_\text{char}$$\downarrow$ & Max LCS$_\text{word}$$\downarrow$ & Max ACS$\downarrow$ & Win rate$\uparrow$ & Rel. & Faith. & Hallu. & F1  \\ 
\midrule
\multirow{5}{*}{NewsSpan} & Vanilla & 542 & 126 & 157 & 27.2\% & {3.0} & {2.2} & {2.3} & {47.9\%}  \\ 
\cmidrule(l){2-10}  
& SysPrompt & 542 & 126 & 153 & 33.0\% & \textbf{2.9} & \textbf{2.3} & \textbf{2.3} & 44.2\%  \\ 
& MemFree & \underline{73} & \underline{18} & \underline{91} & 44.7\% & \underline{2.8} & 2.0 & \underline{2.2} & 45.0\% \\ 
& R-CAD & 291 & 57 & 114 & \underline{54.8\%} & 2.6 & 2.0 & 1.8 & \textbf{47.9\%} \\ 
& \method{} (ours) & \textbf{54} & \textbf{11} & \textbf{63} & \textbf{55.7\%} & \textbf{2.9} & \underline{2.1} & {2.1} & \underline{47.8\%} \\ 
\midrule
\multirow{5}{*}{NewsQA} & Vanilla & 314 & 64 & 117 & 26.7\% & {3.5} & {2.8} & {2.9} & {27.7\%} \\ 
\cmidrule(l){2-10}  
& SysPrompt & 575 & 106 & 109 & 33.3\% & \underline{3.3} & \underline{2.6} & \underline{2.7} & \underline{27.4\%} \\ 
& MemFree & \underline{164} & \underline{30} & \underline{88} & 41.5\% & \textbf{3.4} & \textbf{2.7} & \textbf{2.8} & 25.8\%  \\ 
& R-CAD & 218 & 44 & 90 & \textbf{65.3\%} & 2.7 & 2.4 & 2.2 & \textbf{27.7\%} \\ 
& \method{} (ours) & \textbf{50} & \textbf{11} & \textbf{84} & \underline{52.7\%} & \underline{3.3} & 2.5 & 2.5 & \textbf{27.7\%} \\ 
\bottomrule
\end{tabular}
}
\caption{Infringement against ground truth, information quality, and utility results. \method{} outperforms all methods on worse-case infringement and is competitive on average-case win rate, while preserving information quality and utility.} 
\label{tab:metric_results}
\end{table*}

%% file: tabs/abstention.tex
\begin{table*}[ht]
\centering
\resizebox{\linewidth}{!}{
\begin{tabular}{@{}llcccccccc@{}}
\toprule
\multirow{2}{*}{Dataset} & \multirow{2}{*}{Method} & \multicolumn{2}{c}{\textit{\textbf{Infringement (corpus-level)}}$\downarrow$} & \multicolumn{3}{c}{\textit{\textbf{Infringement (against GT)}}$\downarrow$} & \multicolumn{3}{c}{\textit{\textbf{Info Quality}}$\uparrow$}  \\
\cmidrule(l){3-4} \cmidrule(l){5-7} \cmidrule(l){8-10} 
& & $\%R>Q(50)$ & $\%R>Q(100)$ & Max LCS$_\text{char}$ & Max LCS$_\text{word}$ & Max ACS & Rel. & Faith. & Hallu. \\
\midrule
\multirow{2}{*}{NewsSpan} &\method{} & 10.9\% & 0.0\% & {54} & {11} & {63} & \textbf{2.9} & \textbf{2.1} & {2.1} \\
& +Abstention & \textbf{0.0\%} & \textbf{0.0\%} & \textbf{41} & \textbf{10} & {63} & {2.6} & {2.0} & \textbf{2.4} \\
\midrule
\multirow{2}{*}{NewsQA} &\method{} & 5.6\% & 0.1\% & {50} & {11} & {84} & \textbf{3.3} & \textbf{2.5} & {2.5} \\
& +Abstention & \textbf{0.0\%} & \textbf{0.0\%} & \textbf{42} & {11} & {84} & {3.1} & {2.4} & \textbf{2.6} \\

\bottomrule
\end{tabular}
}

\caption{\textbf{Certified risk reduction} can be achieved through abstention at the cost of small info quality drop (\Sref{sec:ablation}).}
\label{tab:abstain}
\end{table*}

%% file: tabs/ablation.tex
\begin{table*}[ht]
\centering
\resizebox{\linewidth}{!}{
\begin{tabular}{@{}llcccccccc@{}}
\toprule
\multirow{2}{*}{Dataset} & \multirow{2}{*}{Method} & \multicolumn{2}{c}{\textit{\textbf{Infringement (corpus-level)}}$\downarrow$} & \multicolumn{3}{c}{\textit{\textbf{Infringement (against GT)}}$\downarrow$} & \multicolumn{3}{c}{\textit{\textbf{Info Quality}}$\uparrow$}  \\
\cmidrule(l){3-4} \cmidrule(l){5-7} \cmidrule(l){8-10} 
& & $\%R>Q(50)$ & $\%R>Q(100)$ & Max LCS$_\text{char}$ & Max LCS$_\text{word}$ & Max ACS & Rel. & Faith. & Hallu. \\
\midrule
\multirow{2}{*}{Newsspan} &\method{} & \textbf{10.9\%} & \textbf{0.0\%} & \textbf{54} & {11} & {63} & {2.9} & {2.1} & {2.1} \\
& -Quote guidance & {16.8\%} & {0.1\%} & {58} & {11} & {63} & {2.9} & \textbf{2.2} & {2.1} \\
\midrule
\multirow{2}{*}{NewsQA} &\method{} & \textbf{5.6\%} & 0.1\% & \textbf{50} & \textbf{11} & {84} & {3.3} & {2.5} & {2.5} \\
& -Quote guidance & {12.1\%} & \textbf{0.0\%} & {74} & {16} & {84} & {3.3} & {2.5} & {2.5} \\

\bottomrule
\end{tabular}
}

\caption{Ablations shows that quote guidance during rewriting step of \method{} is crucial for effective risk reduction (\Sref{sec:ablation}).}
\label{tab:ablate}
\end{table*}